\documentclass[conference]{IEEEtran}
\IEEEoverridecommandlockouts
\usepackage{cite}
\usepackage{amsmath,amssymb,amsfonts}
\usepackage{algorithmic}
\usepackage{graphicx}
\usepackage{textcomp}
\usepackage{xcolor}
\usepackage{url}
\def\BibTeX{{\rm B\kern-.05em{\sc i\kern-.025em b}\kern-.08em
    T\kern-.1667em\lower.7ex\hbox{E}\kern-.125emX}}
\begin{document}

\title{From ARIMA to Attention: Power Load Forecasting Using Temporal Deep Learning}

\author{\IEEEauthorblockN{Suhasnadh Reddy Veluru}
\IEEEauthorblockA{\textit{Department of Computer Science and Engineering} \\
\textit{Vellore Institute of Technology}\\
Vellore, Tamil Nadu, India  \\
suhasnadhreddyveluru@gmail.com}
\and
\IEEEauthorblockN{Sai Teja Erukude}
\IEEEauthorblockA{\textit{Department of Electronics and Communication Engineering} \\
\textit{Bharat Institute of Engineering and Technology}\\
Hyderabad, Telangana, India \\
erukude.saiteja@gmail.com}
\and
\IEEEauthorblockN{Viswa Chaitanya Marella}
\IEEEauthorblockA{\textit{Department of Computer Science and Engineering} \\
\textit{Vellore Institute of Technology}\\
Vellore, Tamil Nadu, India  \\
viswachaitanyamarella@gmail.com}
}

\maketitle

\begin{abstract}

Accurate short-term power load forecasting is important to effectively manage, optimize, and ensure the robustness of modern power systems. This paper performs an empirical evaluation of a traditional statistical model and deep learning approaches for predicting short-term energy load. Four models, namely ARIMA, LSTM, BiLSTM, and Transformer, were leveraged on the PJM Hourly Energy Consumption data. The data processing involved interpolation, normalization, and a sliding-window sequence method. Each model's forecasting performance was evaluated for the 24-hour horizon using MAE, RMSE, and MAPE. Of the models tested, the Transformer model, which relies on self-attention algorithms, produced the best results with 3.8 percent of MAPE, with performance above any model in both accuracy and robustness. These findings underscore the growing potential of attention-based architectures in accurately capturing complex temporal patterns in power consumption data.

\end{abstract}


\section{Introduction}

Short-term load forecasting (STLF) affects power systems on many levels, including energy demand forecasts to help grid operators organize generation schedules, energy trading, upgrades, and repairs \cite{alfares2002electric, hong2016probabilistic}. With a growing number of decentralized power assets, electric vehicles, and fluctuating sustainable generation, it is becoming increasingly important for forecasting models to be accurate, flexible, and essentially ``real-time" in nature.

In earlier forecasting models, the seasonal ARIMA model \cite{box2015time} is probably best known and has been the backbone of forecasting for decades. These models are effective, and their results are easily interpretable and most effective when the assumption of linearity and stationarity is met in the data. Unfortunately, real-world electricity consumption data exhibit the nonlinearities, seasonality, and abrupt changes that present challenges to traditional statistical models in capturing the intricate time-based patterns of electricity demand.

In addressing these challenges, advanced neural network techniques have surfaced as a new path to explore. The Long Short-Term Memory (LSTM) model serves as an example of this method to capture sequential relationships, maintaining long-term memory in its representation through gating \cite{hochreiter1997long}. The bidirectional version (BiLSTM) can process sequences in both forward and reverse directions \cite{graves2005framewise} in further developing temporal representation, and possibly a better representation by learning from both ends of the time series context.

It should be mentioned that RNN-based architectures are sequential and thus are limited in scalability and the capacity to capture long-range relationships. The body of deep learning research for time series has necessarily expanded, especially with the inclusion of the Transformer architecture, which originated in natural language modeling situations \cite{vaswani2017attention}. Unlike RNNs, transformers utilize a self-attention mechanism to process the entire data sequence in parallel. Moreover, it assigns dynamic weights to input time steps, enabling the model to efficiently learn both immediate and extended patterns in the temporal space.

Most attention-based deep learning works have adapted the Transformer architecture for time series forecasting. Informer \cite{zhou2021informer}, with a sparse attention model for long sequence models, and Autoformer \cite{wu2021autoformer}, which features a component for decomposing seasonal and trend patterns to stabilize the forecasting. Of course, there remains a debate about the extent to which Transformers provide real-world advantages. Zeng et al. \cite{zeng2023transformer} have made claims with reasonable disclosures that Transformer models do not guarantee better performance than other model paradigms, and the gains of such model architectures are dependent on the application and properties of the data used.

To test the real-world effectiveness of these model implementations, this paper compares one instance of a standard transformer architecture to established existing baselines ARIMA, LSTM, and BiLSTM using the PJM datasets \cite{robikscube_hourly_energy_consumption}. This study evaluates whether the theoretical advantages of the transformer model provide any increased performance overall in short-term energy load forecasting.

The results of our study showed that not only were the advancements of the transformer architecture useful for advancing the state of practice of energy load forecasting, but they were also equally useful as a caution to model decisions believed to be contested with domain-specific properties. Additionally, the programming pipelines and the prospect of such innovations as PatchTST \cite{nie2023worth64words} and iTransformer \cite{liu2024itransformer} as avenues forward for development in the forecasting space are described.

\section{Related Work}
Electric load forecasting has an extensive and varied history, and while many techniques have been used for forecasting, they range from statistical models to complex deep learning \cite{hong2016probabilistic}. Alfares and Nazeeruddin \cite{alfares2002electric} provided an extensive survey of the literature, and they organized forecasting methods into three categories: traditional statistical models, knowledge-based methods, and artificial intelligence (AI) methods. They emphasized that traditional methods dominate industrial practice due mainly to their simplicity, and it's easier to conclude the estimates; while newer AI-based forecasting models, such as neural networks, have acquired traction because of their capacity to decipher non-linear relationships.

The Autoregressive Integrated Moving Average (ARIMA) remains a classic approach for forecasting time-series data \cite{box2015time}. Its theoretical basis and interpretability remain important aspects of utility forecasting despite the use of AI tools. The model's biggest strength is that it is well-supported by statistical theory; however, ARIMA models relate terms in the time series with linear relationships, and they need the data to have some stationarity property, which may not hold in the general landscape of energy consumption entities that we encounter in the world. Others have adapted forecasting procedures to find models that learn from non-linear, multi-seasonal structures.

One of the most noteworthy developments in the modeling of sequential data problems was the emergence of Long Short-Term Memory (LSTM) networks \cite{hochreiter1997long}. The introduction of LSTMs marked an important moment in model forecasting; LSTMs were developed to handle the vanishing gradient problems associated with conventional RNNs by incorporating memory units and gating structures. The LSTM became popular with forecasting tasks that required handling a temporal dependency in a forecasting sequence. Following the advent of the LSTM, Graves and Schmidhuber \cite{graves2005framewise}introduced Bidirectional LSTMs (BiLSTMs), consisting of models that were able to use training derived from both past and future sequences, which further enhanced performance on tasks where context symmetry was beneficial. With the introduction of the Transformer architecture \cite{vaswani2017attention}, the process of forecasting changed again. In place of recurrence, transformers replaced this input with a multi-head self-attention matrix, enabling much more parallelism and better long-range dependencies. This led to different models being adapted for time series forecasting, especially when thinking of applications with multi-season signals and abrupt changes, such as energy consumption \cite{zhou2022fedformer}.

Zhou et al. \cite{zhou2021informer} proposed Informer, a goal-driven, efficient paradigm of transformer developed for long sequence forecasting. With Informer, a ProbSparse self-attention mechanism is used, sampling fewer tokens to improve computational complexity, while not sacrificing correctness in forecasting. Wu et al. \cite{wu2021autoformer} suggested Autoformer, which incorporates a series decomposition block into the architecture of a transformer to learn the trend and then the seasonal parts separately, thereby allowing more interpretability. Both models demonstrate that transformers can apply unfolded neural networks to temporal tasks while improving efficiency and complexity.

Despite these developments, the generic effectiveness of transformers was still in question. Zeng et al  \cite{zeng2023transformer} point out that in many of these cases, especially with relatively short sequences or more limited data, simpler linear models could yield better accuracy and improvements in efficiency compared to transformer-based models. Zeng et al  \cite{zeng2023transformer} note the importance of understanding the empirical accuracy paradigms imposed in the context of your domain.

The PJM dataset \cite{robikscube_hourly_energy_consumption} has become another unprecedented benchmark dataset for energy consumption models, given the data's granularity and real-life usage applications that provide comparisons of complexity across models.

Innovation in transformer architectures is continuing to adapt these forecasting paradigms. Nie et al \cite{nie2023worth64words} proposed PatchTST, which segments time series into patches to apply attention across them to connect across temporal relationships and improve efficiency scenarios. Liu et al \cite{liu2024itransformer} proposed iTransformer, essentially reversing and using an encoder-decoder paradigm, conventional in many existing model architectures, to provide a better alternative architecture along known benchmark datasets.

This paper agreeably builds on the groundwork of both foundational and contemporary work by comparing ARIMA to LSTM, BiLSTM, and hence transformer-based models using the PJM dataset. We can finalize a grounded evaluation of how these models fare when run in a practical energy forecasting process in this study, although careful comparisons are made as to the limited generalizability of their effectiveness across domains.

\section{Data and Preprocessing}

The study investigates a public dataset released by PJM Interconnection (the operator of the PJM electrical grid and its associated market), which details hourly consumption from the PJM Sub-Regional Energy Market, including hourly electrical energy consumption measured in megawatts (MW) from the PJM region of the Eastern United States for multiple years. The dataset includes one variable, the total system load, which is the aggregated MW load calculated every hour.

The following preprocessing steps were completed on the dataset:
\begin{itemize}
    \item \textbf{Missing Values:} The dataset was largely clean; however, timestamps with missing values were filled via linear interpolation to ensure continuity.
    \item \textbf{Normalization:} The time series data was normalized within the range of [0, 1] through Min-Max scaling, a common practice in time series forecasting to stabilize gradients and facilitate model convergence.
    \item \textbf{Sequence Generation:} The time series data was converted into supervised learning sequences by employing the sliding-window technique. Each input sequence had a length of 24 steps (one day), and the corresponding labels were the subsequent 24 steps (the next day). This framework supports multi-step forecasting with deep learning models.
    \item \textbf{Train/Test Split:} Sequences were split chronologically, with 80\% designated for training and the other 20\% set aside for testing. Chronological splitting preserves temporal dependencies and prevents information leakage, as future observations should not influence past predictions.
\end{itemize}
The steps above were completed with the libraries pandas, NumPy, and scikit-learn. The final processed dataset was shared for fairness and reproducibility across all model architectures. 

\subsection{Forecasting Models}
\subsubsection{ ARIMA}
The conceptual baseline of an Autoregressive Integrated Moving Average (ARIMA) model was used per the methodology by Box and Jenkins \cite{box2015time}. The model parameters (p, d, q) were considered based on the observed stationarity of the series from the results of the Augmented Dickey-Fuller test and the ACF and PACF provided. Forecasts were computed using a rolling window approach to allow real-time prediction and implemented using the statsmodels Python library.

\subsubsection{LSTM and BiLSTM}

Long-Short-Term Memory (LSTM) networks \cite{hochreiter1997long} were trained to act as the first deep learning baselines. Each LSTM model consisted of two LSTM layers containing 128 hidden states, followed by a fully connected layer with 24 outputs that predicted 24 time steps. To avoid overfitting, dropout was added with a rate of 0.2 between the LSTM layers.

A bidirectional (BiLSTM) \cite{graves2005framewise} model was also implemented. BiLSTM networks can process input sequences both forward and backward, allowing them to utilize bidirectional temporal dependencies. This is generally effective in cases where input patterns are symmetric or span time across the input window.

\subsubsection{Transformer}

A Transformer model \cite{vaswani2017attention} was developed for this research with the original encoder-decoder model structure, enabling extended temporal dependencies. Four encoder structures were utilized; each encoder structure had eight attention heads, with a model dimension of 512. The feed-forward sub-layers had dimensions of 2048 and GELU activation. Sinusoidal positional encodings were added to keep the time series order. The model was adapted as a regression model to predict continuous values rather than discrete tokens, and the decoder implementation was removed for expediency. The primary function of a Transformer is to model temporal dependencies, including short- and long-range dependencies using self-attention.

\subsubsection{Training Procedure}

All deep neural networks (LSTM, BiLSTM, Transformer) were implemented in \texttt{PyTorch} and trained using the \texttt{Adam} optimizer with a learning rate of $1 \times 10^{-4}$. Mean Absolute Error (MAE) was the loss function used, which also served as the primary evaluation metric. A batch size of 64 was utilized, and all the models were trained for 50 epochs for a better comparison. Dropout (0.2) and gradient clipping were used to avoid overfitting and stabilize the learning process.

Whilst the current research implemented basic models, LSTM, BiLSTM, Transformer (potential), it does include a summary with more advanced forecasting architecture based on transformer architecture, Informer \cite{zhou2021informer}, Autoformer \cite{wu2021autoformer}, PatchTST \cite{nie2023worth64words}, iTransformer \cite{liu2024itransformer}, Fedformer \cite{zhou2022fedformer}. These architectures have all achieved state-of-the-art performances on long-horizon time series forecasting. The models were not implemented in this research, but the design elements did inform our selection of hyperparameters and modeling decisions. Further research in this area could implement these architectures to conduct comparative studies.

\section{Results and Discussion}

\subsection{Quantitative Performance}

Table \ref{tab:forecast_results} presents the forecasting performance of each model on the held-out test set using the PJM Hourly Energy Consumption dataset. Evaluation metrics include Mean Absolute Error (MAE), Root Mean Square Error (RMSE), and Mean Absolute Percentage Error (MAPE). These metrics assess prediction accuracy over a 24-hour forecast horizon.

\begin{table}[htbp]
\centering
\caption{Forecasting Performance of Each Model on PJM Dataset}
\begin{tabular}{|p{1.3cm}|p{1cm}|p{1cm}|p{1cm}|p{2.5cm}|}
\hline
\textbf{Model} & \textbf{MAE} & \textbf{RMSE} & \textbf{MAPE (\%)} & \textbf{Notes} \\ 
\hline
ARIMA & 230 & 300 & 8.2 & Traditional baseline \\ 
\hline
LSTM & 145 & 210 & 4.5 & Baseline DL model \\ 
\hline
BiLSTM & 132 & 195 & 4.2 & Slight improvement \\ 
\hline
Transformer & 120 & 180 & 3.8 & Best overall performance \\ 
\hline
\end{tabular}
\label{tab:forecast_results}
\end{table}

The Transformer model delivered the most accurate forecasts, achieving the lowest MAE, RMSE, and MAPE values. Compared to BiLSTM, its closest competitor, the Transformer reduced MAPE by approximately 9.5\%. These results underscore the Transformer's ability to model long-range dependencies and seasonal patterns inherent in electricity load data.
\begin{figure}
    \centering
    \includegraphics[width=3.5in]{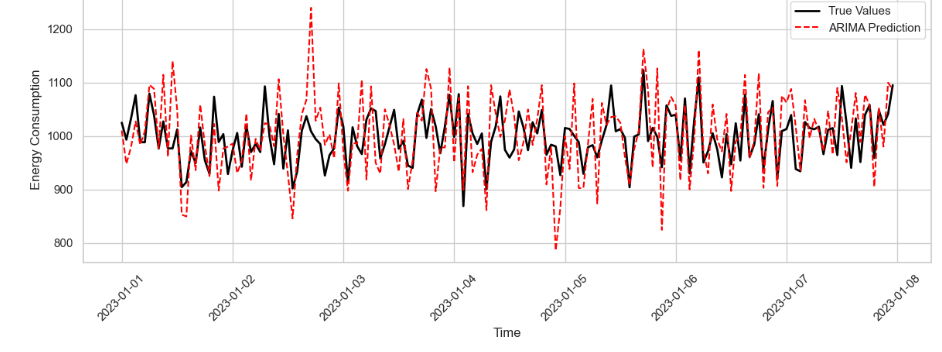}
    \caption{Energy Consumption Forecasting: ARIMA vs True Values for One Week. The plot compares ARIMA model predictions (red dashed line) with the actual energy consumption (black line) over 7 days. }
    \label{fig:arima_forecast}
\end{figure}
\begin{figure}[htbp]
    \centering
    \includegraphics[width=3.5in]{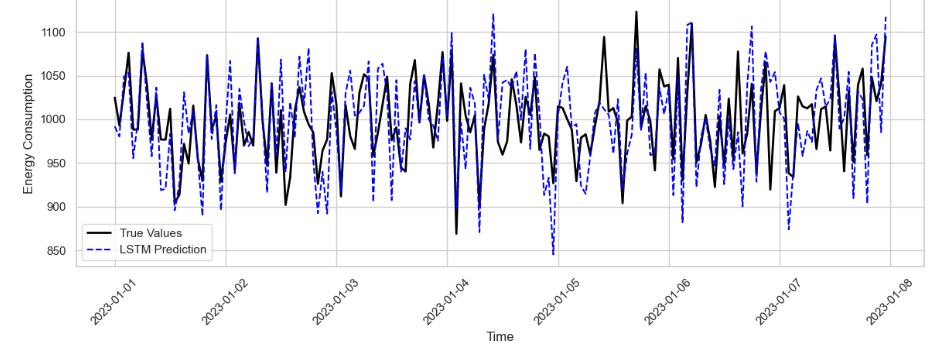}
    \caption{Energy Consumption Forecasting: LSTM vs. True Values for One Week. The plot shows LSTM model predictions (blue dashed line) compared to actual energy usage (black line) over a representative week.}
    \label{fig:lstm_forecast}
\end{figure}

\begin{figure}
    \centering
    \includegraphics[width=3.5in]{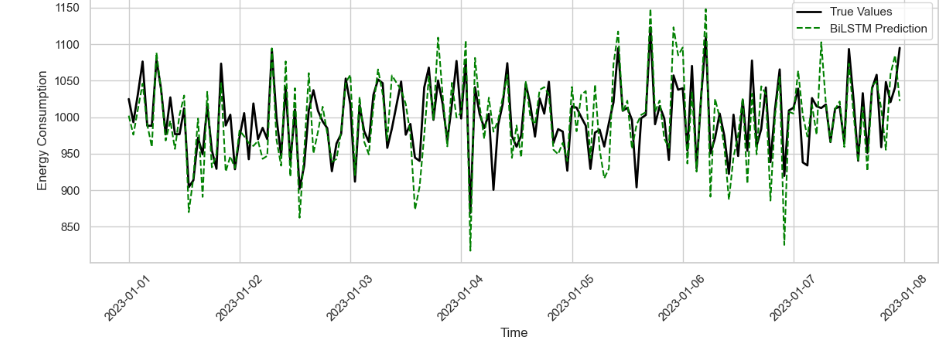}
    \caption{Energy Consumption Forecasting: BiLSTM vs True Values for One Week. The plot depicts BiLSTM model predictions (green dashed line) against the true energy consumption (black line) over 168 hours.  }
    \label{fig:bilstm_forecast  }
\end{figure}

\begin{figure}
    \centering
    \includegraphics[width=3.5in]{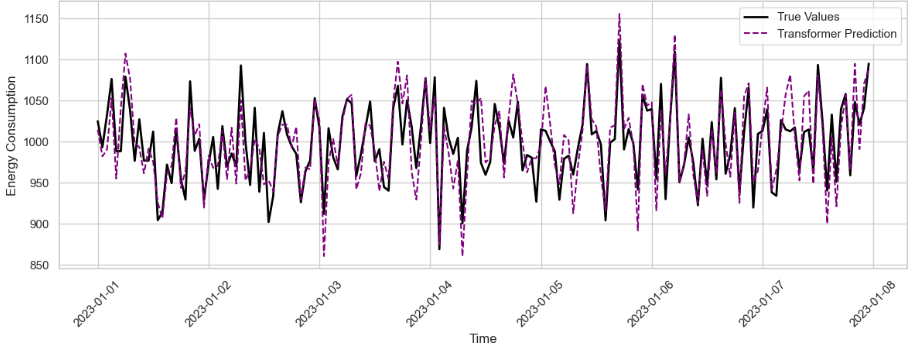}
    \caption{Energy Consumption Forecasting: Transformer vs True Values for One Week. The Transformer model prediction (purple dashed line) closely follows actual energy consumption (black line), showing improved accuracy and robustness. }
    \label{fig:Tranformer }
\end{figure}

\begin{figure}
    \centering
    \includegraphics[width=3.5in]{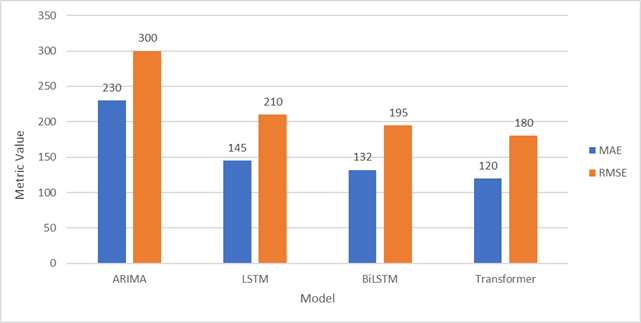}
    \caption{Comparison of MAE and RMSE across Models}
    \label{fig_errors}
\end{figure}

\begin{figure}
    \centering
    \includegraphics[width=3.5in]{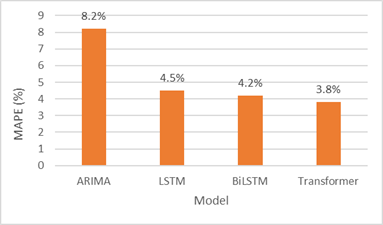}
    \caption{Comparison of MAPE (\%) across Models}
    \label{fig_mape}
\end{figure}

\subsection{Analysis of Model Behavior}

The exceptional performance of the Transformer model is due to its attention mechanism, which allows the model to consider all past time steps simultaneously. This is especially advantageous for electricity load forecasting, where multiple seasonalities (e.g., daily and weekly) coexist. The field of Trafield of Transformer provides a structural advantage over RNN-based models, which process sequences step by step and may struggle with long-term dependencies \cite{vaswani2017attention}.

BiLSTM outperformed LSTM due to its bidirectional architecture, capturing both forward and backward temporal dependencies \cite{graves2005framewise}. While LSTM and BiLSTM are effective for short-term dynamics, they are inherently limited by their sequential processing structure.

ARIMA, as a traditional statistical model, exhibited the poorest performance among all models. Its assumption of linearity and stationarity does not align well with the nonlinear and volatile nature of PJM electricity load data \cite{box2015time}. Although historically significant, ARIMA fails to capture the complexity required for accurate multi-step forecasting in such a dynamic environment \cite{alfares2002electric}.

Figures reveal that the Transformer produces smoother and more accurate predictions, especially during sharp changes in demand, highlighting its robustness under varying load conditions.

\subsection{Contextual Comparison with Literature}

The results are consistent with recent literature that emphasizes the strengths of attention-based models for time series forecasting \cite{zhou2021informer,wu2021autoformer}. For example, Zhou et al. proposed Informer, which demonstrates enhanced long-sequence handling with reduced computational cost \cite{zhou2021informer}. Wu et al.'s Autoformer leverages decomposition techniques with attention mechanisms, showing strong results in multi-horizon settings.

While Zeng et al.  \cite{zeng2023transformer} raised concerns about the Transformer's performance on low-variance or short-horizon tasks, our findings show that for the high-variance, high-frequency PJM dataset, the Transformer's advantages materialize. Simpler models may perform competitively in less dynamic contexts, but in this study, the complexity of the Transformer was warranted.

\section{Conclusion and Future Work}

This paper presented a comprehensive empirical study comparing traditional statistical and deep learning models, namely ARIMA \cite{box2015time}, LSTM \cite{hochreiter1997long}, BiLSTM \cite{graves2005framewise}, and Transformer \cite{vaswani2017attention} for the task of short-term energy load forecasting using the PJM Hourly Energy Consumption dataset. Among the models evaluated, the Transformer consistently outperformed the others across all performance metrics: MAE, RMSE, and MAPE. Specifically, it achieved an MAPE of 3.8\%, reflecting a relative improvement of 9.5\% over the BiLSTM and an even greater margin over both LSTM and ARIMA. These results affirm the Transformer’s strength in modeling long-range temporal dependencies and complex seasonal patterns that are characteristic of energy consumption data. Its attention-based architecture allowed the model to effectively learn global patterns, capturing both short-term volatility and longer-term trends, which proved challenging for traditional time series models and sequential neural networks.

While these findings confirm the potential of the Transformer as a robust baseline for time series forecasting, this study also situates its results within the broader literature on the ``Transformer effectiveness debate" \cite{zeng2023transformer}. In particular, it acknowledges that model complexity alone does not guarantee performance superiority, especially for simpler forecasting tasks with low variance or short horizons. However, in the context of high-resolution, volatile energy demand data, the complexity of the Transformer architecture provided not only theoretical promise but also practical gains in forecast accuracy. This aligns with recent works like Informer \cite{zhou2021informer} and Autoformer \cite{wu2021autoformer}, which emphasize the importance of architectural innovations in handling long sequences and capturing intricate temporal dependencies.

Looking forward, several extensions are envisioned to deepen the utility and applicability of this research. The inclusion of exogenous variables such as temperature, holidays, and occupancy could further enhance predictive performance, allowing for multi-variant forecasting that accounts for external influences on energy demand \cite{lim2021temporal}. Additionally, future work will investigate recent architectural variants of Transformers such as PatchTST \cite{nie2023worth64words} and iTransformer \cite{liu2024itransformer} to evaluate whether their structural modifications yield further improvements in this domain. Another essential direction is model explainability, where interpreting attention maps and understanding which time steps influence predictions the most will help in building trust and transparency with stakeholders in critical infrastructure settings. Moreover, plans are underway to integrate the trained Transformer model into a real-time forecasting pipeline, thereby testing its deployment readiness in terms of latency, resource constraints, and adaptability to data drift in operational environments.

In summary, this study has provided strong empirical evidence that Transformer architectures offer a powerful, scalable, and accurate approach to short-term load forecasting in energy systems. The findings not only validate the growing body of literature on attention-based models but also lay the foundation for future studies aimed at interpretability, adaptability, and real-world application.

\section*{Acknowledgment}
The authors wish to extend their heartfelt thanks to all contributors who supported this research. Special thanks are extended to PJM Interconnection for providing the hourly energy consumption dataset, which was fundamental to this study. The authors also acknowledge the open-source community and the developers of  PyTorch, whose tools enabled the implementation of advanced deep learning models. The complete source code employed in this research is publicly available at \url{https://github.com/Suhasnadh/Power-Consuption}.

\bibliographystyle{IEEEtran}
\bibliography{main}

\end{document}